\documentclass[10pt,twocolumn,letterpaper]{article}
\usepackage[pagenumbers]{cvpr}   % [pagenumbers] = arXiv / preprint version with author names and page numbers
\usepackage{graphicx,booktabs,multirow,amsmath,amssymb,siunitx,xcolor,microtype}
\definecolor{cvprblue}{rgb}{0.21,0.49,0.74}   % as in the CVPR 2026 kit's main.tex
\usepackage[pagebackref,breaklinks,colorlinks,allcolors=cvprblue]{hyperref}
\usepackage{xurl}                % long URLs in the bibliography break between characters instead of stretching the line
\usepackage{bm}
\usepackage{placeins}
\newcommand{\ours}{HaiRuo}       % the whole system (brand name, as Qwen)
\newcommand{\fm}{\texorpdfstring{HaiRuo\protect\nobreakdash-1B}{HaiRuo-1B}}     % label-free pretrained backbone (Zhuangzi, "Autumn Floods": the god of the North Sea); no line break at the hyphen, safe in headings and bookmarks
\newcommand{\dec}{XuanJi}     % object decoder (armillary rings)

\newcommand{\best}[1]{\textbf{#1}}
\newcommand\blfootnote[1]{%
  \begingroup
  \renewcommand\thefootnote{}\NoHyper\footnote{\normalfont #1}\endNoHyper%
  \addtocounter{footnote}{-1}%
  \endgroup\ignorespaces
}
\def\confName{CVPR}
\def\confYear{2027}

\title{\ours: Forecasting Ocean Eddies as Fields and Objects with a Satellite-Altimetry Foundation Model}

\author{%
Xukun Qin$^{1,2}$\qquad Haochen Sun$^{1}$\qquad Yuyang Tian$^{2,3}$\qquad Ge Chen$^{1,*}$\\[3pt]
{\small $^{1}$Ocean University of China\quad $^{2}$Shanghai Artificial Intelligence Laboratory\quad $^{3}$University of Science and Technology of China}%
}

\makeatletter
\g@addto@macro\@maketitle{%
  \begin{center}\vspace*{-18pt}
  \includegraphics[width=0.86\textwidth]{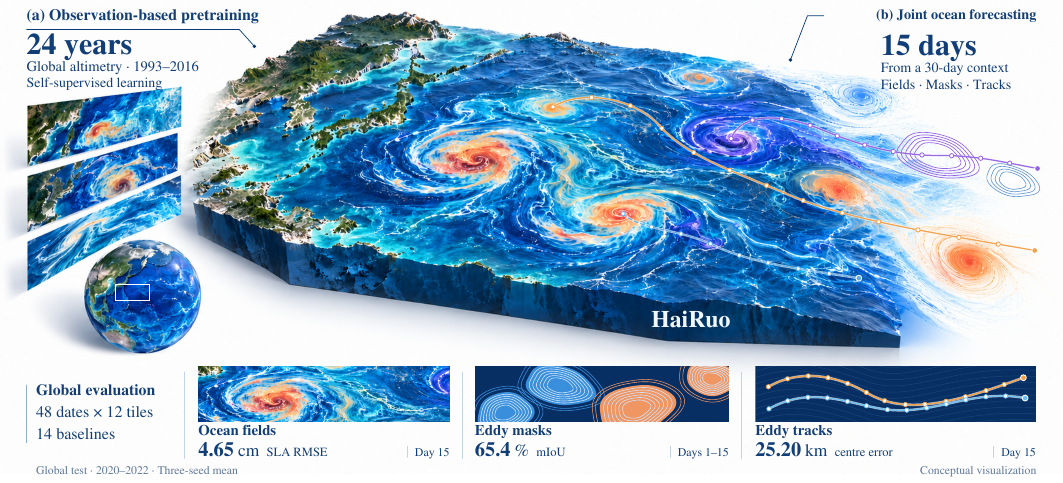}\par\vspace{2pt}
  \captionof{figure}{\textbf{\ours{} forecasts ocean fields, eddy masks and eddy trajectories for 15 days in one forward pass.}
  (a) \fm{} is pretrained without labels on 24 years of altimetry and then fine-tuned under \dec{}; (b) \dec{} reads it out as fields, masks and every
  eddy's trajectory. Bottom: \ours{} on the global 2020--2022 evaluation, 48 dates $\times$ 12 tiles, three-seed means; the fourteen
  baselines are in Table~\ref{tab:global}.}\label{fig:teaser}
  \end{center}\vspace{0pt}}
\g@addto@macro\abstract{\blfootnote{$^*$Corresponding author.}}
\makeatother

\makeatletter
\global\@namedef{r@app:protocol}{{A}{S1}{}{}{}}
\global\@namedef{r@app:training}{{B}{S1}{}{}{}}
\global\@namedef{r@app:algo}{{C}{S2}{}{}{}}
\global\@namedef{r@tab:pretrain_abl}{{4}{S3}{}{}{}}
\global\@namedef{r@app:fourregion}{{D}{S4}{}{}{}}
\global\@namedef{r@app:enso}{{E}{S5}{}{}{}}
\global\@namedef{r@tab:enso_global_full}{{8}{S6}{}{}{}}
\global\@namedef{r@fig:enso_eke}{{6}{S7}{}{}{}}
\global\@namedef{r@app:physics}{{F}{S8}{}{}{}}
\global\@namedef{r@app:baselines}{{G}{S9}{}{}{}}
\global\@namedef{r@app:wenhai}{{H}{S9}{}{}{}}
\global\@namedef{r@tab:global_systems}{{12}{S9}{}{}{}}
\global\@namedef{r@app:more}{{I}{S9}{}{}{}}
\global\@namedef{r@tab:components_full}{{14}{S12}{}{}{}}
\global\@namedef{r@tab:samedecoder}{{15}{S13}{}{}{}}
\global\@namedef{r@fig:lead_errors}{{7}{S15}{}{}{}}
\global\@namedef{r@fig:case_grid_3rows}{{8}{S16}{}{}{}}
\global\@namedef{r@fig:xuanji_detail}{{9}{S17}{}{}{}}
\global\@namedef{r@fig:same_decoder}{{10}{S17}{}{}{}}
\makeatother
\begin{document}
\maketitle
% ===== begin sections/abstract =====
% Float anchor: Figure 2 (overview) is anchored here, before the abstract, so that LaTeX meets it on page 1 and places it at the top of
% page 2 (a figure* never lands on the page where it is anchored). Anchoring it before \section{Introduction} fails when the abstract
% fills page 1 exactly, because the heading, and the anchor with it, then moves to page 2 and the figure to page 3.
\begin{figure*}[t]\centering
\includegraphics[width=\textwidth]{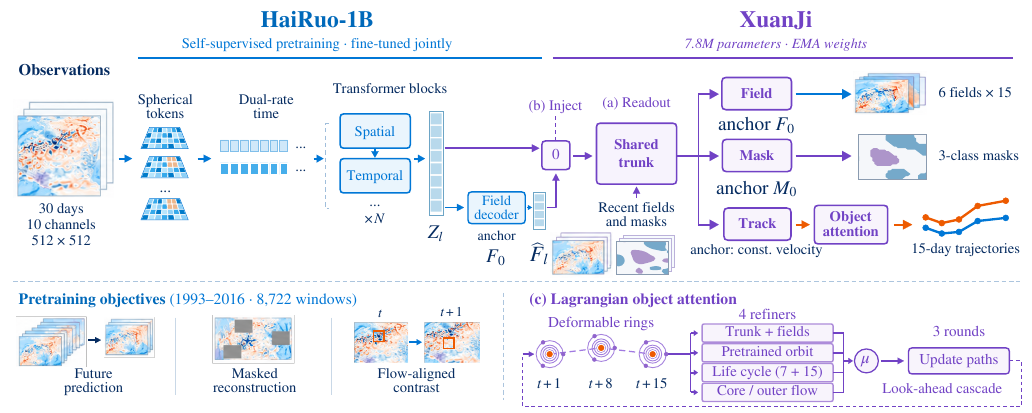}
\caption{\textbf{\dec{} reads the \fm{} forecast into fields, masks and trajectories.} \fm{} turns 30 days of altimetry into
spherical tokens, compresses time at two rates and applies spatio-temporal blocks. Its per-lead features $Z_l$ and its own forecast
$\hat F_l$ are injected (b) into \dec{}'s multi-scale spatial stem and proposal heads (a), anchored at persistence and constant velocity; \dec{} object attention (c) samples the forecast and the backbone along every
predicted path and refines all 15 leads in three look-ahead rounds, averaged ($\mu$, $\omega_j{=}\tfrac14$) over four refiners.}
\label{fig:overview}
\end{figure*}
\begin{abstract}
Mesoscale eddies carry most of the ocean's kinetic energy and move heat, salt and nutrients across basins, yet where an eddy will be two
weeks ahead is still read off ocean models with post-hoc tracking heuristics. We introduce \textbf{\ours{}}, which forecasts eddies
\textbf{as fields and as objects at once}: from 30 days of satellite sea-level maps, one forward pass returns 15 days of sea level and
geostrophic velocity, daily eddy masks and the trajectory of every eddy alive today. \ours{} pairs two components with a clear division of
labour: \textbf{(1) \fm{}}, an open-weight \textbf{1.05-billion-parameter} foundation model pretrained without labels on \textbf{24 years} of global
satellite-altimetry observations, which supplies the 15-day physical fields and masks ($+0.14$ field skill and $+1.1$ mIoU points under the same decoder) and further lowers 15-day centre error by $1.2$\,km frozen ($1.35$\,km over training from scratch); and \textbf{(2) \dec{}}, a Lagrangian object decoder that samples the evolving forecast in each eddy's co-moving polar frame and refines the 15-day path across three look-ahead rounds, removing $2.5$\,km on \fm{} ($2.9$\,km with joint fine-tuning) and $2.5$--$3.3$\,km via its object attention across six published backbones. Under one pre-registered global evaluation over the whole ocean, \ours{} improves on \textbf{fourteen} methods on every endpoint (\textbf{25.2 vs.\ 30.6\,km} 15-day centre error, \textbf{\textminus18\,\%}), holding its margin outside the labelled regions and through both phases of the El Ni\~no--Southern Oscillation (ENSO), while reproducing westward drift, 50--60$^\circ$S eastward reversal and poleward cyclone--anticyclone deflection. Over ten days through one fixed tracker, \ours{}'s forecast field cuts the eddy-centre error of WenHai and XiHe by \textbf{37--43\,\%} ($-2.8$\,km vs.\ day 0 while they degrade $+4.3$--$8.7$\,km), and its object trajectories cut pipeline error by 77--78\,\% globally.
\end{abstract}
% ===== end sections/abstract =====
% ===== begin sections/intro =====
% Float anchor: Figure 2 (overview) is defined at the top of sections/abstract.tex, i.e. while page 1 is still being built, so that it
% lands at the top of page 2 even when the abstract fills page 1 completely and this heading starts page 2.
\section{Introduction}
\label{sec:intro}
Where will this eddy be in two weeks? Mesoscale eddies, coherent vortices 50--300\,km across that live for weeks to months, carry most of the
ocean's kinetic energy and move heat, salt and nutrients across basins~\cite{chelton2011,ferrari2009ocean}. Satellites have mapped them every day for three decades, and atlases detect and track the
eddies the maps resolve~\cite{pegliasco2022,faghmous2015}. Forecasting them is a different matter. Run a state-of-the-art AI ocean
forecaster~\cite{wenhai2025,xihe2024} and track the eddies in its forecast with a fixed tracker, and they land farther from their true
positions than if the forecaster's own initial state had been held fixed (Section~\ref{sec:system}).

The reason is structural. These systems forecast fields, learn from reanalysis, and leave eddies to a tracker applied afterwards; nothing in
them follows an eddy. Seen from computer vision, eddy forecasting is coupled field-and-object forecasting on continuous spherical fields, where eddies are coherent vortices defined by gradients of the evolving flow itself: given 30
daily maps, predict the next 15, the daily cyclonic and anticyclonic masks, and the path of every eddy alive today. We find that
\textbf{the object belongs in the decoder}: a forecaster should learn the ocean's dynamics at scale from what satellites observe, and then
read eddies out as objects that follow and re-sample its own evolving forecast.

We present \textbf{\ours{}} (Figure~\ref{fig:teaser}). Existing AI ocean models~\cite{wenhai2025,xihe2024,langya2024,glonet2025,fuxiocean2025} are reanalysis state emulators rather than open, label-free observation foundation backbones for downstream decoders. Our backbone, \fm{}, fills this gap: a \textbf{1.05-billion-parameter} foundation model pretrained without labels on \textbf{24 years} of
global satellite-altimetry observations with a two-stage spherical $0.5^\circ$ stem, dual-rate 30-day memory, eddy-focused chirality-gated attention, multi-axis masked reconstruction and a geostrophic flow-advected contrast. Every readout tested benefits from \fm{}, and we pair it with \dec{}, a Lagrangian object decoder that returns 15-day fields, daily masks and every eddy's trajectory via \emph{\dec{} object attention}: unlike static-map motion queries~\cite{shi2022mtr,zhou2023qcnet}, occupancy grids~\cite{hu2021fiery,hu2023uniad} or point-vortex advection, \dec{} re-samples the model's own continuous evolving 15-day forecast and backbone along each eddy's predicted path in a co-moving polar frame (mean intensity, self-advection dipole and strain quadrupole modes), attends across all 15 leads, and averages four refiners over three look-ahead rounds.

Our contributions are:
(i) \textbf{\fm{}}, an open-weight 1.05B ocean foundation model pretrained without labels on 24 years of global altimetry \emph{observations},
forecasting 15-day sea level and geostrophic velocity in one pass and lifting every readout tested: under \dec{}, it raises field skill from 0.63 to 0.77, adds 1.1 mIoU points and lowers D15 by 1.2\,km frozen on Dev48 (1.35\,km over training the same 1B architecture from scratch, and 0.6--1.4\,km over six video backbones), while even with plain base heads frozen \fm{} (25.8\,km) beats all 14 baselines and self-supervised PredRNN-V2 (30.4\,km frozen, 26.6\,km fine-tuned);
(ii) \textbf{\dec{}}, a plug-and-play co-moving polar object decoder that samples azimuthal Fourier modes of the evolving forecast along each eddy's predicted orbit and refines all 15 leads jointly across three look-ahead rounds and four refiners, removing \textbf{2.5\,km} of D15 on \fm{} on Dev48 (\textbf{2.9\,km} with joint fine-tuning; 0.56\,km on Dev48 and \textbf{0.87\,km} globally from multi-expert refinement) and \textbf{2.5--3.3\,km} via two-round object attention across six published backbones;
(iii) a pre-registered global evaluation (Table~\ref{tab:global}, Figure~\ref{fig:global_map}) in which \ours{} improves on \textbf{fourteen methods on every endpoint}, lowering D15 by \textbf{18\,\%} against the strongest and by 17--18\,\% outside the labelled regions and in both ENSO phases, while reproducing latitudinal westward drift (and 50--60$^\circ$S eastward reversal) and poleward cyclone--anticyclone deflection; and
(iv) a system comparison through one fixed tracker over days 1--10, in which \ours{}'s forecast field cuts the
eddy-centre error of WenHai and XiHe by \textbf{37--43\,\%} field for field on Test48 (improving by 2.8\,km over its held-fixed day-0 state while they degrade by 4.3--8.7\,km), and direct object trajectories cut pipeline error by 77--78\,\% over the whole ocean.
% ===== end sections/intro =====
% ===== begin sections/related =====
\section{Related work}
\label{sec:related}
\paragraph{Spatio-temporal prediction.} Recurrent predictors (ConvLSTM~\cite{shi2015}, PredRNN-V2~\cite{wang2022predrnn},
PhyDNet~\cite{leguen2020phydnet}, SwinLSTM~\cite{tang2023swinlstm}), recurrence-free models (SimVP~\cite{gao2022simvp}, TAU~\cite{tan2023tau},
SimVPv2~\cite{tan2025simvpv2}), transformers (Earthformer~\cite{gao2022earthformer}, PredFormer~\cite{tang2024predformer}) and neural operators
(FNO~\cite{li2021fno}) are the standard tools for video and weather-like forecasting. They treat every pixel or patch alike.

\paragraph{Eddy forecasting.} Eddy atlases detect eddies as closed sea-surface-height contours (earlier, by Okubo--Weiss~\cite{chelton2007global}) and link them in time~\cite{chelton2011,pegliasco2022,mason2014pyeddy}; EddyNet~\cite{lguensat2018}
learns detection. Trajectory forecasters model each track from its own attributes via recurrent~\cite{mesogru2021}, transformer~\cite{eddytransformer2024}, dual-graph~\cite{dualgraph2025} and LSTM/Informer~\cite{eddylsin2025} networks, or via physics-fused~\cite{eddytpnet2024} and global~\cite{mediumrange2024} models without public code, seeing
at most local field descriptors. \ours{} forecasts fields, masks and tracks jointly, reading its own forecast around every moving eddy.

\paragraph{Ocean forecasters and foundation models.} Learned ocean forecasters (XiHe~\cite{xihe2024}, WenHai~\cite{wenhai2025},
LangYa~\cite{langya2024}, GLONET~\cite{glonet2025}, FuXi-Ocean~\cite{fuxiocean2025}) train on reanalysis from assimilated three-dimensional
states and forecast state variables rather than eddy objects~\cite{oceanbench2025}. Altimetry interpolators~\cite{beauchamp2023fourdvarnet} and climate emulators~\cite{dheeshjith2025samudra} likewise do not forecast eddies as objects, and open foundation backbones pretrained on global ocean observations remain scarce.
Earth-system foundation models (Aurora~\cite{aurora2025}, Prithvi WxC~\cite{prithviwxc2024},
ClimaX~\cite{climax2023}) pretrain on atmospheric reanalyses and climate simulations with one- to two-step histories
and $\ge1^\circ$ tokens. \fm{} pretrains directly on altimetry observations with a 30-day context and $0.5^\circ$ eddy-resolving tokens, matched to eddies that drift a few kilometres per day with a 50\,km median
radius.

\paragraph{Object-centric decoding.} Deformable attention~\cite{zhu2021deformable} and slot attention~\cite{locatello2020slot} gather
evidence where the object is, and object-centric video predictors roll slots forward in time~\cite{wu2023slotformer}. In motion
forecasting and planning, MTR~\cite{shi2022mtr} and QCNet~\cite{zhou2023qcnet} refine trajectories with static map features gathered around their own
proposals, joint detection--forecasting models refine along estimated past motion~\cite{masar2026}, FIERY~\cite{hu2021fiery} and
UniAD~\cite{hu2023uniad} predict future occupancy grids alongside instance trajectories, and classical vortex-track models advect point vortices in a steering flow. In contrast, \dec{} object attention samples the model's own continuous multi-channel forecast of an evolving field on deformable co-moving
polar rings where the eddy is \emph{going}, decomposes each ring into azimuthal Fourier modes ($m{=}0,1,2$), and re-estimates all 15 leads jointly.
% ===== end sections/related =====
% ===== begin sections/method =====
\section{Method}
\label{sec:method}

\subsection{Problem and overview}
\label{sec:problem}
The input is a window of $T{=}30$ daily altimetry maps $X\in\mathbb{R}^{30\times10\times H\times W}$ on the $0.125^\circ$ grid ($H{=}W{=}512$,
a $64^\circ\times64^\circ$ tile) with ten channels: sea-level anomaly (SLA), absolute dynamic topography (ADT), zonal and meridional geostrophic velocity
anomalies and absolute velocities, SLA mapping error, latitude, longitude and Coriolis parameter, plus an ocean mask
and per-day quality map (Appendix~\ref{app:protocol}). At readout, the decoder also reads the observed masks and eddy tracks of the last seven days up to $t_0$. For the
next $L{=}15$ days the model predicts (i) the six dynamic fields $\hat F_l$, (ii) a three-class mask $\hat M_l$ (background, cyclonic,
anticyclonic) and (iii) the centre $\hat p_{e,l}$ of every eddy $e$ that exists on the last observed day $t_0$. Labels for (ii) and (iii)
come from a global satellite-altimetry eddy atlas (Section~\ref{sec:setup}); (i) needs none.

\ours{} has two parts (Figure~\ref{fig:overview}): the backbone \fm{} (Section~\ref{sec:backbone}), pretrained without labels on 24 years of
global altimetry and fine-tuned jointly with the decoder at a learning rate of $3\times10^{-5}$, and the object decoder \dec{} (Section~\ref{sec:decoder}). Every head is zero-initialised: before training
the forecast is persistence, the mask is the last observed mask and every trajectory is constant-velocity extrapolation.

% Float anchors (all on page 3, in this order): the double-column floats fill the page tops in sequence, Figure 3 (global map) on
% page 4, Figure 4 (case study) on page 5, Table 1 (global evaluation) on page 6 right above Section 4, and Figure 5 (skill and reliability) on page 7.
\begin{figure*}[t]\centering
\includegraphics[width=0.92\textwidth]{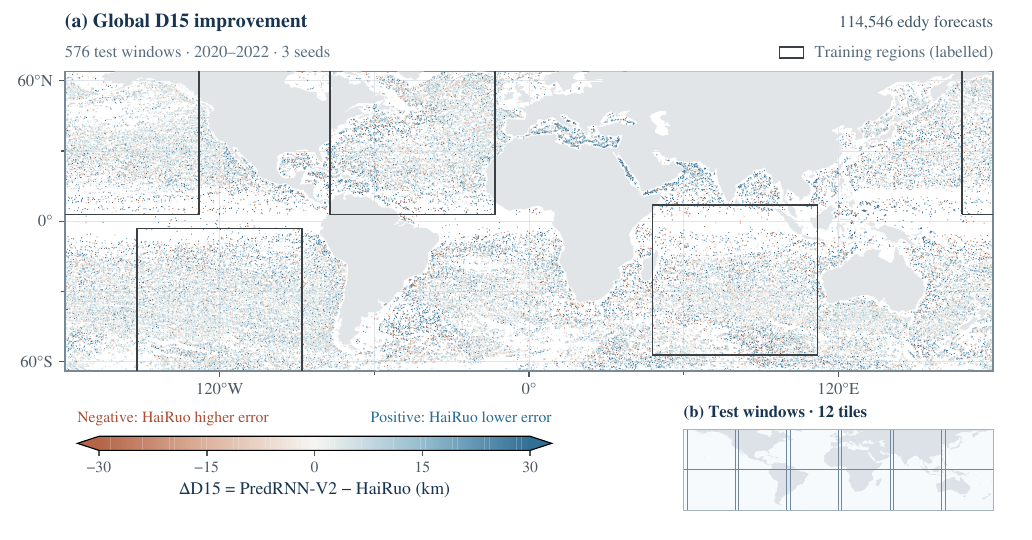}
\caption{\textbf{\ours{} lowers the 15-day centre error across the whole ocean, including the regions never seen in labelled training.}
(a) Day-0 position of every evaluated eddy, coloured by the 15-day centre error (D15) of PredRNN-V2 minus that of \ours{} (three-seed mean;
blue favours \ours{}), with the four labelled training regions outlined. (b) The 12 evaluation tiles.}
\label{fig:global_map}
\end{figure*}
\begin{figure*}[t]\centering
\includegraphics[width=0.75\textwidth]{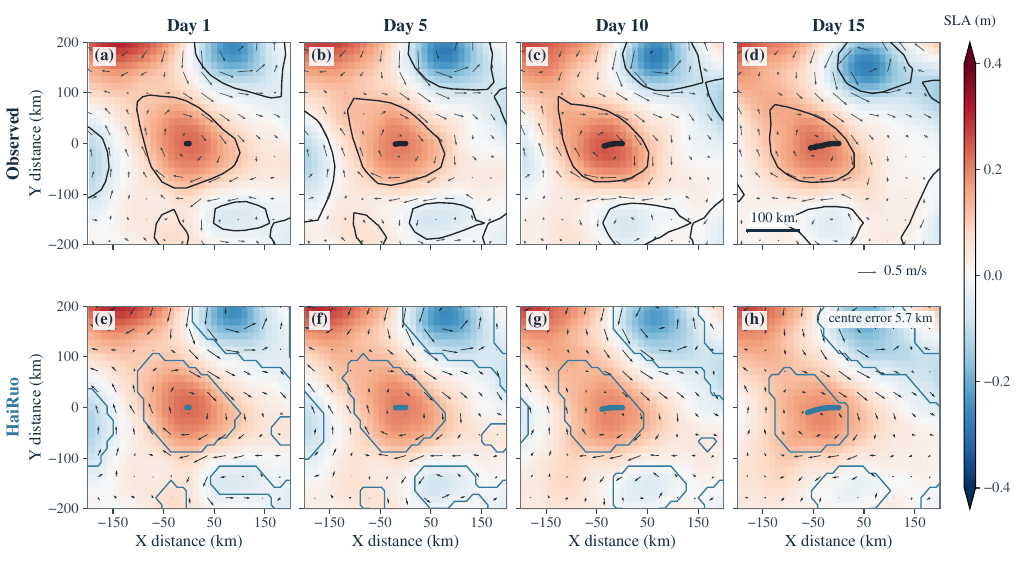}
\caption{\textbf{\ours{} follows an anticyclone through its own forecast.} A window of the global evaluation issued on 12 August 2022 in the
Kuroshio Extension: observed (top) and forecast (bottom, seed 42) sea-level anomaly with geostrophic velocity (arrows), eddy contours
(atlas above, forecast mask below) and the centre track (dots) at leads 1, 5, 10 and 15, in a fixed 400\,km frame centred on the eddy's
day-0 position. The eddy drifts 56\,km in 15 days; the forecast centre ends 5.7\,km from the observed one (PredRNN-V2: 8.4\,km;
Figure~\ref{fig:case_grid_3rows}); over the global evaluation the median eddy forecast of \ours{} ends 17\,km from the observed centre
(PredRNN-V2: 22\,km).}
\label{fig:case_grid}
\end{figure*}
% ===== begin tables/global =====
\begin{table*}[t]\centering\footnotesize\setlength{\tabcolsep}{3.2pt}
\caption{Global evaluation on the independent 2020--2022 period: 48 dates on 12 tiles covering all longitudes between 64$^\circ$S and 64$^\circ$N, every model scored once. Training regions: the four regions whose labelled windows are used for training; other regions: the rest of the ocean, never seen in labelled training. Three-seed means (seed s.d.\ of D15: 0.03\,km for \ours{}, 0.11--0.80\,km for the field baselines, 0.02 and 0.80\,km for LSTM+Informer and Track Transformer); trajectory-only models have no masks or fields. Field: field skill; L15: lead 15. Bold: best in the column. $\Delta$: \ours{} minus the method in the other regions, with its 95\,\% interval and the relative change.}
\label{tab:global}
\begin{tabular}{@{}lcccccccc@{}}
\toprule
 & \multicolumn{3}{c}{D15 (km) $\downarrow$} & Centre $\downarrow$ & mIoU $\uparrow$ & Field $\uparrow$ & SLA RMSE $\downarrow$ & $\Delta$D15, other regions \\
\cmidrule(lr){2-4}
Method & Global & \shortstack{Training\\regions} & \shortstack{Other\\regions} & (km) & (\%) & & L15 (cm) & km [CI], rel. \\
\midrule
\best{\ours{} (ours)} & \best{25.20} & \best{22.74} & \best{27.64} & \best{14.03} & \best{65.40} & \best{0.736} & \best{4.65} & -- \\
\midrule
PredRNN-V2 (2023) & 30.58 & 27.53 & 33.61 & 17.01 & 63.31 & 0.556 & 5.39 & $-5.97$ [$-6.16$, $-5.79$], $-17.8$\,\% \\
PhyDNet (2020) & 30.59 & 27.54 & 33.63 & 16.89 & 63.50 & 0.549 & 5.40 & $-5.99$ [$-6.20$, $-5.79$], $-17.8$\,\% \\
TAU (2023) & 30.64 & 27.73 & 33.52 & 16.70 & 63.50 & 0.578 & 5.46 & $-5.88$ [$-6.07$, $-5.70$], $-17.6$\,\% \\
SimVPv2/gSTA (2025) & 30.85 & 27.91 & 33.77 & 16.81 & 63.21 & 0.588 & 5.42 & $-6.13$ [$-6.31$, $-5.95$], $-18.2$\,\% \\
SimVP (2022) & 30.96 & 28.11 & 33.80 & 16.85 & 62.15 & 0.520 & 5.80 & $-6.16$ [$-6.32$, $-6.01$], $-18.2$\,\% \\
U-Net (2015) & 31.00 & 28.09 & 33.88 & 16.94 & 59.93 & 0.309 & 7.56 & $-6.24$ [$-6.43$, $-6.06$], $-18.4$\,\% \\
ConvLSTM (2015) & 31.19 & 28.24 & 34.13 & 17.07 & 63.28 & 0.553 & 5.49 & $-6.49$ [$-6.70$, $-6.30$], $-19.0$\,\% \\
EddyNet (2018) & 32.88 & 30.27 & 35.47 & 17.73 & 61.98 & 0.438 & 6.04 & $-7.83$ [$-8.04$, $-7.63$], $-22.1$\,\% \\
PredFormer (2026) & 33.20 & 30.56 & 35.82 & 17.97 & 60.57 & 0.411 & 6.06 & $-8.18$ [$-8.43$, $-7.96$], $-22.8$\,\% \\
Earthformer (2022) & 33.41 & 30.38 & 36.43 & 18.05 & 61.16 & 0.459 & 5.75 & $-8.79$ [$-8.99$, $-8.60$], $-24.1$\,\% \\
LSTM+Informer (2025, tracks) & 34.05 & 31.21 & 36.89 & 18.29 & {--} & {--} & {--} & $-9.25$ [$-9.69$, $-8.90$], $-25.1$\,\% \\
SwinLSTM-B (2023) & 34.28 & 31.53 & 37.02 & 18.63 & 58.96 & 0.130 & 6.60 & $-9.38$ [$-9.62$, $-9.15$], $-25.3$\,\% \\
FNO-2d (2021) & 34.90 & 32.14 & 37.65 & 19.01 & 53.83 & 0.251 & 6.48 & $-10.01$ [$-10.26$, $-9.77$], $-26.6$\,\% \\
Track Transformer (2024, tracks) & 36.66 & 34.05 & 39.24 & 19.58 & {--} & {--} & {--} & $-11.60$ [$-11.82$, $-11.38$], $-29.6$\,\% \\
\bottomrule
\end{tabular}
\end{table*}
% ===== end tables/global =====
\begin{figure*}[t]\centering
\includegraphics[width=0.9\textwidth]{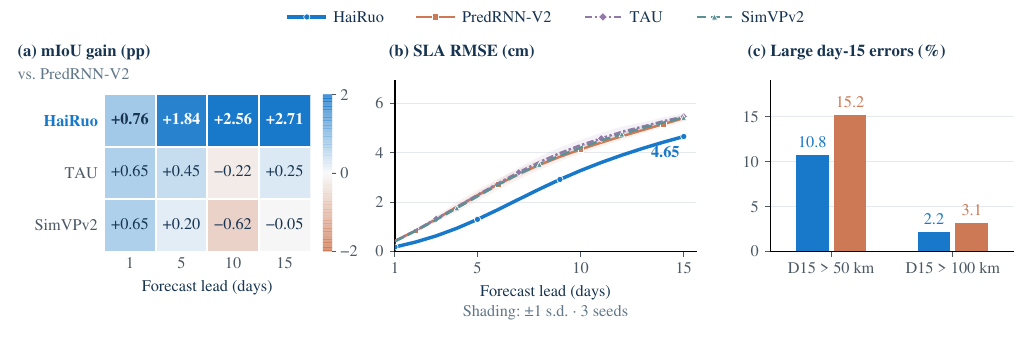}
\caption{\textbf{Skill by lead and reliability on the global evaluation.} (a) Three-class mIoU of the lead-$l$ mask relative to PredRNN-V2
(difference of three-seed means, percentage points). (b) Sea-level-anomaly RMSE by lead (three-seed means; shading one standard deviation
across seeds). (c) Share of the 114\,546 evaluated eddies whose day-15 centre error, averaged over three seeds, exceeds 50 and 100\,km.}
\label{fig:global_metrics}
\end{figure*}

\subsection{\fm{}: pretrained foundation backbone}
\label{sec:backbone}
A two-stage spherical stem embeds each daily map into fine tokens at $0.5^\circ$ (patch 4), so a 50\,km-radius eddy spans one to two tokens without aliasing; two spherical stride-2
convolutions build a three-level pyramid at $0.5^\circ$, $1^\circ$ and $2^\circ$ (depths 6/10/20), with a learned quality gate damping poorly observed swaths. To capture month-scale evolution without quadratic cost, the 30-day history is compressed at two rates: the last
seven days stay explicit and the preceding 23 are distilled by per-token softmax routing into eight chronological memory slots (15 steps in total), keeping all 30 days on a differentiable path. The
three levels stack 36 eddy-focused spatio-temporal blocks: spatial attention combines validity-masked shifted windows with soft eddy-vorticity focus, core--ring deformable sampling and chirality-gated projections, while temporal attention applies physical day-age-aligned QKV routing across the 15 steps. A bank of 240 sparse eddy route-memory tokens and global latents~\cite{jaegle2021perceiver} bridges the pyramid to a multi-scale field
decoder that queries each lead, fuses the three levels through longitude-periodic residual blocks, and outputs all 15 daily residuals for the six fields in one pass,
\begin{equation}
\hat F_l = F_0 + \delta_l\odot r_l,
\label{eq:contract}
\end{equation}
where $F_0$ is the last observed field and $\delta_l$ the training RMS of the $l$-day change; zero-initialised $r_l$ starts every
model at persistence (the 200M ablation scale uses $1^\circ$ tokens; Appendix~\ref{app:training}).

\subsection{Self-supervised pretraining on altimetry}
\label{sec:pretraining}
Pretraining visits every daily window of 1993--2016 (8\,722 windows) and draws an area-uniform random $512\times512$ tile on the global
grid, so \fm{} learns dynamics across the whole ocean. Three self-supervised objectives are optimised jointly: (i) 15-lead future prediction via an area-weighted Huber loss on
the $\delta_l$-normalised residuals of all six fields; (ii) swath-aware multi-axis masked reconstruction~\cite{he2022mae} (40\,\% of spatial tokens, $\le15\,\%$ of full days, $\le20\,\%$ of channels) via cross-attention to the compressed timeline; and (iii) a geostrophic
flow-advected InfoNCE contrast~\cite{oord2018cpc} pairing each day-$t$ token at $p_t$ with its day-$(t{+}1)$ counterpart bilinearly sampled at $\tilde p_{t+1}{=}\mathrm{Exp}_{p_t}(\bm{u}_g(p_t)\Delta t)$ along the observed spherical geostrophic velocity $\bm{u}_g$
(Appendix~\ref{app:training}). Pretraining runs from random initialisation for 93 epochs (12\,648 updates at batch size 64).

\subsection{\dec{}: object decoder}
\label{sec:decoder}
\paragraph{Forecast-conditioned spatial stem and initial proposal.}\label{sec:readout}
A multi-scale residual convolutional stem (the shared trunk of Figure~\ref{fig:overview}a) reads the last seven days of altimetry, seven days of past eddy masks, the ocean mask and \fm{}'s 15-lead forecast, while \fm{}'s per-lead features enter the stem through zero-initialised $1\times1$ projections, producing a spatial feature map per lead at stride 4. Three zero-initialised heads read these features: a residual field head under
(\ref{eq:contract}), a mask head refined from the last observed mask, and a proposal trajectory head that integrates on the sphere a velocity
correction from a gated recurrent unit (GRU) over each eddy's seven-day track and a residual velocity sampled at five
points around the constant-velocity proposal to form the initial path $\hat p^{(0)}_{1:L}$. Without \fm{} and object attention, this stem and its three heads serve as the shared Eulerian readout for all field baselines.

\paragraph{\dec{} object attention.}\label{sec:lagrangian}
For an eddy with radius $R$ and position estimate $p_l$ at lead $l$, we sample features on five rings ($r/R\in\{0.5,1,1.5,2,3\}$) of 16
azimuths centred at $p_l$ in the eddy's Lagrangian (co-moving) frame (Figure~\ref{fig:xuanji_detail}; Appendix~\ref{app:algo}), and reduce each ring to its azimuthal Fourier modes $m=0,1,2$: the mean, the dipole
that encodes the asymmetry driving self-advection, and the quadrupole that encodes strain. The rings read the spatial-stem features, the six forecast
fields and, directly, \fm{}: its per-lead features and its own forecast, together with the step velocity and acceleration of the current path and the geostrophic velocity implied by the dipole modes of sea level
(the \emph{pretrained-orbit readout}: the backbone read along the eddy's predicted orbit). With the GRU state, the eddy metadata and a lead encoding, the modes form one token per lead; a two-layer
transformer attends across the 15 lead tokens, so the correction at lead 10 can depend on what the eddy meets at lead 5, and a
zero-initialised layer outputs a velocity correction per lead. Sampling and correction run for $K{=}3$ rounds ($k{=}0,\dots,K{-}1$) at the updated positions with
learned ring deformations (Appendix~\ref{app:algo}). With $\mathcal{S}$ the ring sampling, $\phi_j$ the reduction to modes and tokens for refiner $j$
(next paragraph), $u^{(k)}_j{=}\phi_j(\mathcal{S}(\hat p^{(k)}_{1:L}))$ its lead tokens in round $k$, $g_j$ its zero-initialised output layer
and $D^{(k)}$ the displacement accumulated over the rounds so far ($D^{(0)}{=}0$),
round $k$ updates the whole path at once,
\begin{equation}
D^{(k+1)}_{l} = D^{(k)}_{l} + \textstyle\sum_{l'\le l}\sum_{j=1}^{4}\omega_j\, g_j\!\big(\mathrm{Attn}^{j}_{1:L}[u^{(k)}_j]\big)_{l'},
\label{eq:loa}
\end{equation}
and $\hat p^{(k+1)}_{1:L} = \hat p^{(0)}_{1:L} \oplus \tfrac{K}{k+1} D^{(k+1)}_{1:L}$, where $\oplus$ integrates a displacement on the
sphere from the initial path, $\mathrm{Attn}^{j}_{1:L}$ is refiner $j$'s transformer over its lead tokens, the sum over $l'$ turns
daily velocities into a displacement and $\omega_j{=}\tfrac14$ at inference. The factor $K/(k{+}1)$ is the \emph{look-ahead cascade}:
every round samples where the eddy will be at the end of its path, carries its lead states and ring deformations into the next, and the
last round ($k{+}1{=}K$) applies the displacement as it is.

\paragraph{Multi-expert look-ahead refinement.} The four refiners $g_1,\dots,g_4$ of (\ref{eq:loa}) read complementary evidence along the eddy's trajectory: the spatial-stem and forecast rings ($g_1$), the pretrained
\fm{} orbit ($g_2$), the eddy's 22-day past-and-future life cycle ($g_3$), and the geostrophic contrast between
inner core and outer rings ($g_4$). In training, a randomly drawn primary refiner dominates each eddy's correction ($\omega_{\text{primary}}{=}0.76$, others $0.08$) to prevent co-adaptation; at inference the four are averaged ($\omega_j{=}\tfrac14$; Appendix~\ref{app:algo}).

\paragraph{Training.} The objective sums a Huber loss on the fields, focal--Dice, boundary and soft-Jaccard losses on the masks, and a
smooth-$\ell_1$ loss on positions and daily displacements. Training visits 1\,024 labelled windows eight times with AdamW and
cosine decay and evaluates the weights at the fixed final step; \dec{} is evaluated with an exponential moving average (EMA) of its
weights over the last three quarters of training. \dec{} (7.85M parameters) and \fm{} (1.06B total) train jointly in 10--12\,h on one GPU (Appendices~\ref{app:algo},~\ref{app:more}).
% ===== end sections/method =====
% ===== begin sections/experiments =====
\section{Experiments}
\label{sec:experiments}

\subsection{Setup}
\label{sec:setup}
\paragraph{Data and protocol.} Inputs are daily $1/8^\circ$ delayed-time DUACS altimetry maps~\cite{taburet2019duacs,cmems2026quid} and
META4.0 eddy atlas labels~\cite{gamot2026meta4} (Appendix~\ref{app:protocol}; operational forecasting would use near-real-time streams), split into 1993--2016 (pretraining and labelled training), 2017--2019 (Dev48) and 2020--2022 (Test48).
Labelled windows come from four fixed $512\times512$ regions (North Atlantic, North Pacific, South Pacific, Indian Ocean); the main evaluation
covers the whole ocean between 64$^\circ$S and 64$^\circ$N. All models follow one pre-registered protocol: the same 1\,024 labelled windows
and training budget, the same base readout heads for the field methods (Section~\ref{sec:readout}), three seeds and the fixed final checkpoint. The test period is scored
once per model globally and on Test48 (Appendices~\ref{app:protocol},~\ref{app:fourregion}), with 95\,\% paired bootstrap intervals.

\paragraph{Metrics.} \emph{D15}: great-circle error (km) of the predicted centre at lead 15 for every atlas eddy alive at $t_0$ and at
lead 15 (Appendix~\ref{app:protocol});
\emph{centre}: the same error over leads 1--15; \emph{mIoU}: three-class mean intersection over union of the daily masks; \emph{field skill}:
$1-\mathrm{MSE}/\mathrm{MSE}_{\rm persistence}$ of the six fields (persistence 0, perfect 1; time-constant MDT cancels in $\hat F_l{-}F_l$ and $F_0{-}F_l$, so ADT and absolute velocities share identical errors and skill with SLA and velocity anomalies).

\paragraph{Baselines.}\label{sec:baselines} We re-implement fourteen methods as complete systems under the same labelled-training budget, the twelve field
methods with the same base readout heads (Appendix~\ref{app:baselines}):
ConvLSTM~\cite{shi2015}, PredRNN-V2~\cite{wang2022predrnn}, PhyDNet~\cite{leguen2020phydnet}, SwinLSTM-B~\cite{tang2023swinlstm},
SimVP~\cite{gao2022simvp}, TAU~\cite{tan2023tau}, SimVPv2/gSTA~\cite{tan2025simvpv2}, Earthformer~\cite{gao2022earthformer},
PredFormer~\cite{tang2024predformer}, FNO-2d~\cite{li2021fno}, EddyNet~\cite{lguensat2018}, a U-Net~\cite{ronneberger2015unet}, and two trajectory-only models that see the
eddy histories but no field (LSTM+Informer~\cite{eddylsin2025}, Track Transformer~\cite{eddytransformer2024}).

\subsection{Comparison with state-of-the-art methods}
\label{sec:main}\label{sec:global}
Table~\ref{tab:global} reports the independent 2020--2022 test over the whole ocean (Figure~\ref{fig:global_map}), split into the
four training regions and all other regions. \ours{} leads on every endpoint, inside and outside the training regions (Appendix~\ref{app:more}). Over the whole ocean it brings D15
to 25.2\,km (18\,\% below PredRNN-V2, every paired interval excluding zero), field skill to 0.74 and 15-day sea-level error to 4.7\,cm, against 0.59 and
5.4\,cm for the best baseline on each. In Figure~\ref{fig:case_grid} the eddy keeps its closed contour through day 15;
Figure~\ref{fig:global_map} maps the D15 gain eddy by eddy, and
Figure~\ref{fig:global_metrics}a,b follows the mask and sea-level gains by lead. Across the 114\,546 global eddies, the share with D15 above 50\,km falls from
15.2\,\% for PredRNN-V2 to 10.8\,\% (Figure~\ref{fig:global_metrics}c). On Test48, \ours{} likewise leads all baselines on foreground eddy IoU, 1-pixel contour boundary F1, unconditioned all-track recall and short-lived eddy accuracy (Appendix~\ref{app:more}).

Outside the labelled training regions, \ours{} stays 18\,\% ahead of the strongest baseline (TAU), leads all 14 methods on masks, fields and D15 (every paired interval excluding zero), and shows the smallest absolute regional gap (4.9\,km vs.\ 5.2--6.1\,km). Under the shared base heads, the strongest recurrent and recurrence-free predictors all lie within 1\,km of a U-Net, whereas \fm{} pretraining and
\dec{} remove over 5\,km (and still lead when baselines receive \dec{} object attention, 22-epoch global pretraining, or all 8\,722 windows; Section~\ref{sec:attribution}, Appendix~\ref{app:more}).

\subsection{Forecasting through ENSO}
\label{sec:enso}
% ===== begin tables/enso_global_main =====
\begin{table}[t]\centering\footnotesize\setlength{\tabcolsep}{2.5pt}
\caption{Accuracy through ENSO: global evaluation in the other regions (2020--2022), windows grouped by the ENSO phase of their anchor
date (Jan--Mar 2020 El Ni\~no; 17 Dev48 El Ni\~no dates: 23.34 vs.\ 27.29\,km, $-14.5$\,\%; 9 RONI dates: 23.18 vs.\ 27.56\,km, Appendix~\ref{app:enso}). \ours{} against the strongest of the 14 baselines in each phase (bold: best; all methods: Table~\ref{tab:enso_global_full}).}
\label{tab:enso_global}
\begin{tabular}{@{}lcccc@{}}
\toprule
 & \multicolumn{2}{c}{El Ni\~no (4 dates)} & \multicolumn{2}{c}{La Ni\~na (34 dates)} \\
\cmidrule(lr){2-3}\cmidrule(lr){4-5}
 & \ours{} & best baseline & \ours{} & best baseline \\
\midrule
D15 (km) & \best{28.24} & 34.12 (TAU) & \best{27.28} & 33.20 (TAU) \\
\quad gap & \multicolumn{2}{c}{$-5.88$ ($-17.2$\,\%)} & \multicolumn{2}{c}{$-5.91$ ($-17.8$\,\%)} \\
Centre (km) & \best{15.66} & 18.61 (TAU) & \best{15.16} & 18.03 (TAU) \\
mIoU (\%) & \best{63.70} & 61.64 (TAU) & \best{63.86} & 61.71 (TAU) \\
Field skill & \best{0.714} & 0.556 (gSTA) & \best{0.723} & 0.565 (gSTA) \\
\bottomrule
\end{tabular}
\end{table}
% ===== end tables/enso_global_main =====
\ours{} is trained on 1993--2016 and never sees a climate index. Its test period spans both phases of the El Ni\~no--Southern Oscillation
(ENSO), the dominant mode of interannual ocean variability~\cite{mcphaden2006enso}: the 2019--20 El Ni\~no and 2020--22
La Ni\~na winters from the Oceanic Ni\~no Index (ONI)~\cite{cpc2026oni}.

Table~\ref{tab:enso_global} scores the El Ni\~no and La Ni\~na windows outside the training regions separately. \ours{} keeps its
accuracy through both phases, within 2.2\,\% of its all-phase error, and leads all fourteen methods on every endpoint in both phases,
by 17--18\,\% in D15. The margin holds on Test48 and on the 2018--19 Dev48 El Ni\~no (17 ONI dates: $-14.5$\,\%; 9 Relative ONI~\cite{nws2026roni} dates: $-15.9$\,\%). In the tropical Pacific, the margin over PredRNN-V2 stays at 17\,\% (8.9\,km; Figure~\ref{fig:enso_eke}c, Appendix~\ref{app:enso}):
ENSO modulates eastern-Pacific eddy kinetic energy by 38\,\% without affecting \ours{}'s accuracy or its lead over all 14 methods.

\paragraph{Physical consistency.} \ours{} is given no Rossby-wave or $\beta$-drift law, yet its forecasts
reproduce altimetry propagation statistics~\cite{chelton2011} (Appendix~\ref{app:physics}): across global latitude bins
the 15-day westward drift matches observations ($r{=}0.98$, slope 1.14, reversing to eastward at 50--60$^\circ$S), and unlike constant-velocity extrapolation (which inherits zonal drift but reverses meridional polarity deflection in both hemispheres) cyclones drift poleward relative to anticyclones in both hemispheres, closer to observations than PredRNN-V2. Moreover, geostrophic velocities derived from \ours{}'s predicted SLA on Test48 best match observed currents across all 12 field baselines ($r{=}0.901$, 5.47 vs.\ $\ge 5.91$\,cm/s, $r{\le}0.884$).

\subsection{Ablation study}
\label{sec:attribution}
% ===== begin tables/components_main =====
\begin{table}[t]\centering\footnotesize\setlength{\tabcolsep}{1.8pt}
\caption{Ablation study on Dev48 (D15 in km, mIoU in points). Top: frozen backbones read out by \dec{} ($\Delta$ paired on shared seeds vs.\ no backbone). Middle: representative pretraining ablations (all 14 variants in Table~\ref{tab:pretrain_abl}; $^\dagger$memory masked at readout). Bottom: \dec{} components and joint fine-tuning. 95\,\% intervals in Table~\ref{tab:components_full}. Bold: best backbone, full system.}
\label{tab:components}
\resizebox{\columnwidth}{!}{%
\begin{tabular}{@{}lcccc@{}}
\toprule
 & D15 & $\Delta$D15 & $\Delta$mIoU & $\Delta$field \\
\midrule
\multicolumn{5}{@{}l}{\emph{Foundation backbone scaling (same \dec{} readout, vs.\ no backbone)}} \\
None (decoder alone, 2 seeds) & 25.07 & -- & -- & -- \\
200M, $1^\circ$ tokens (22 ep.) & 25.16 & $+0.04$ & $+0.06$ & $+0.014$ \\
200M, $0.5^\circ$ tokens (22 ep.) & 25.01 & $-0.06$ & $+0.34$ & $+0.059$ \\
1B, $1^\circ$ tokens (97 ep.) & 24.64 & $-0.45$ & $+0.51$ & $+0.095$ \\
\best{Frozen \fm{} (1B, $0.5^\circ$, 93 ep.)} & \best{23.85} & \best{$-1.19$} & \best{$+1.09$} & \best{$+0.141$} \\
\midrule
\multicolumn{5}{@{}l}{\emph{Pretraining ablations (effect of removing each component)}} \\
w/o pretraining (1B from scratch, vs.\ \ours{}) & 24.22 & $+1.35$ & $-1.12$ & $-0.107$ \\
w/o two-stage stem \& $0.5^\circ$ tokens (200M) & 25.15 & $+0.15$ & $-0.25$ & $-0.044$ \\
w/o chirality-gated projection (200M, Tab.~\ref{tab:pretrain_abl}) & 25.32 & $+0.16$ & $+0.02$ & $+0.003$ \\
w/o 23-day memory slots (200M, Tab.~\ref{tab:pretrain_abl})$^\dagger$ & 25.12 & $+0.02$ & $-0.06$ & $-0.010$ \\
\midrule
\multicolumn{5}{@{}l}{\emph{\dec{} object decoder and joint fine-tuning}} \\
\dec{} object attention (2 rounds, vs.\ 25.76) & 23.85 & $-1.91$ & $-0.03$ & $0.000$ \\
\dec{} object attention (2 rounds, $1^\circ$ 1B, vs.\ 26.72) & 24.64 & $-2.08$ & $+0.01$ & $+0.001$ \\
\quad w/o second sampling round & & $+0.28$ & $-0.03$ & $0.000$ \\
\quad w/o cross-lead attention & & $+0.12$ & $0.00$ & $-0.001$ \\
\dec{} 4-expert 3-round look-ahead $+$ EMA (vs.\ 23.85) & 23.29 & $-0.56$ & $+0.21$ & $+0.001$ \\
Fine-tuned \fm{} $+$ \dec{} (\ours{}, vs.\ 23.29) & 22.87 & $-0.42$ & $+0.36$ & $-0.004$ \\
\quad \best{Total \dec{} $+$ joint fine-tuning (vs.\ 25.76)} & \best{22.87} & \best{$-2.89$} & \best{$+0.55$} & $-0.003$ \\
\bottomrule
\end{tabular}}
\end{table}
% ===== end tables/components_main =====
Table~\ref{tab:components} and Table~\ref{tab:samedecoder} separate the roles of the foundation backbone and the object decoder on Dev48: \fm{} supplies the physical fields and masks and lifts every readout tested, while \dec{} refines the trajectories of every backbone it reads (\ours{}: 22.87\,km).

\paragraph{Pretraining (\fm{}).} Under the same \dec{} decoder, frozen \fm{} lifts field skill from 0.63 to 0.77 ($+0.141$), adds 1.09 mIoU points and
removes 1.19\,km of D15 over no backbone (25.07 to 23.85\,km): \textbf{all} of the gain of \ours{} over the decoder alone in fields, \textbf{most} in masks and \textbf{over half} in
trajectories comes from the pretrained representation. Trained from scratch, the identical 1B~$+$~\dec{} model scores 24.22\,km, 66.52\,\% mIoU and 0.662 field skill, losing 1.35\,km D15, 1.12 mIoU and 0.107 field skill to \ours{} (trailing even frozen \fm{} by 0.93\,km). Under the same two-round \dec{} decoder, frozen \fm{} also leads all six published video-prediction backbones on every endpoint: by 0.57--1.44\,km D15, 0.80--1.85 mIoU and 0.074--0.142 field skill, every interval excluding zero (Table~\ref{tab:samedecoder}). Even with plain base heads (no object attention), frozen \fm{} (25.76\,km, 67.10\,\% mIoU, 0.771 field skill) beats all 14 baselines and 22-epoch self-supervised PredRNN-V2 pretrained on the same global altimetry record (30.36\,km / 63.31\,\% / 0.735 frozen; 26.58\,km / 66.84\,\% / 0.739 fine-tuned; Table~\ref{tab:components_full}, Appendix~\ref{app:more}). Across backbone scales, field skill and mIoU grow from 200M ($+0.014$--$0.059$ field skill, $+0.06$--$0.34$ mIoU) to 1B (where D15 also drops by $0.45$\,km at $1^\circ$ and $1.19$\,km at $0.5^\circ$), with the two-stage $0.5^\circ$ stem and eddy-focused spatio-temporal blocks ablated in Table~\ref{tab:pretrain_abl} (Appendix~\ref{app:training}).

\paragraph{\dec{}.} Attached as a plug-in to six published backbones, two-round \dec{} object attention lowers D15 on every one of them by 2.49--3.34\,km while leaving masks and fields unchanged (Figure~\ref{fig:same_decoder}, Table~\ref{tab:samedecoder}). On frozen \fm{}, two-round object attention removes 1.91\,km of D15 (25.76 to 23.85\,km; $-2.08$\,km on the $1^\circ$ 1B backbone), with both multi-round sampling ($+0.28$\,km at one round) and cross-lead attention ($+0.12$\,km when removed) contributing (Table~\ref{tab:components}). Full three-round four-expert look-ahead refinement with EMA removes another 0.56\,km on Dev48 (23.29\,km; $-0.77$\,km on Test48, $-0.87$\,km globally), and joint fine-tuning (\ours{}, 22.87\,km; $-2.89$\,km total) removes another 0.42\,km on Dev48 ($-0.34$\,km on Test48, $-0.21$\,km globally) while adding 0.36--0.78 mIoU points.

\subsection{Comparison with global ocean forecasters}
\label{sec:system}
We evaluate the released global AI ocean forecasters WenHai~\cite{wenhai2025} and XiHe~\cite{xihe2024} (from GLORYS12~\cite{lellouche2021glorys}, WenHai forced by ERA5) on the 48 test dates at two levels (Appendix~\ref{app:wenhai}): (i) \emph{field-for-field through the same fixed tracker} on Test48, \ours{}'s forecast field
scores 29.5\,km, cutting WenHai (47.1\,km) and XiHe (51.6\,km) by 37--43\,\%, landing within 0.52\,km of true altimetry (29.0\,km), and improving by 2.8\,km over its held-fixed day-0 state (32.3\,km) while WenHai and XiHe degrade by 4.3 and 8.7\,km beyond theirs (42.9\,km); (ii) \emph{as objects}, \ours{}'s trajectories score 10.1\,km on Test48 (79--80\,\% lower) and 11.2\,km globally (Figure~\ref{fig:lead_errors}b, Table~\ref{tab:global_systems}), 77--78\,\% below WenHai (48.0\,km) and XiHe (52.0\,km), bypassing the 18.8\,km tracker offset on true altimetry and 10.6\,km day-0 reanalysis offset.
% ===== end sections/experiments =====
% ===== begin sections/conclusion =====
\FloatBarrier
\section{Conclusion}
\label{sec:conclusion}
\ours{} forecasts mesoscale eddies as fields and objects from satellite altimetry. Pretrained without labels on 24 years of global altimetry, the open-weight 1.05B foundation model
\fm{} supplies the physical fields and masks and lifts every readout tested, while \dec{} refines eddy trajectories on every backbone it reads; together they beat all 14 methods on every endpoint of a
pre-registered global protocol, cutting D15 by 18\,\% overall and 17--18\,\% outside the labelled regions and in both ENSO phases, while reproducing westward drift (and 50--60$^\circ$S eastward reversal) and poleward cyclone--anticyclone deflection. Open-weight checkpoints, code, pre-registration record and the project page will be available at \url{https://github.com/xkqin/HaiRuo}.\par
% ===== end sections/conclusion =====
\clearpage
{\small\bibliographystyle{ieeenat_fullname}\bibliography{refs}}
\end{document}